\documentclass[sigconf]{acmart}

\renewcommand\footnotetextcopyrightpermission[1]{} 
\AtBeginDocument{%
  }

\acmConference[Conference]{2025}{Under Review}{}

\makeatletter
\gdef\@copyrightpermission{
}
\makeatother

\usepackage[most]{tcolorbox}
\usepackage[table]{xcolor}
\usepackage[normalem]{ulem}
\usepackage{tabularray}
\usepackage{makecell} 
\usepackage[normalem]{ulem}
\useunder{\uline}{\ul}{}
\usepackage{algorithm}
\usepackage{algorithmic}
\usepackage{multirow}
\usepackage{amsmath} 
\usepackage{enumitem}
\usepackage{balance}
\usepackage{caption} 

\usepackage{array} 
\usepackage{etoolbox} 
\usepackage{graphicx}
\usepackage{booktabs}

\usepackage{tabularx}
\usepackage{xcolor}
\usepackage{geometry}

\begin{document}
\title{Towards Unification of Hallucination Detection and Fact Verification for Large Language Models}

\author{Weihang Su}
\email{swh22@mails.tsinghua.edu.cn}
\affiliation{
DCST, Tsinghua University\\
Beijing  \country{China}
}

\author{Jianming Long}
\affiliation{
DCST, Tsinghua University\\
Beijing  \country{China}
}

\author{Changyue Wang}
\affiliation{
DCST, Tsinghua University\\
Beijing  \country{China}
}

\author{Shiyu Lin}
\affiliation{
DCST, Tsinghua University\\
Beijing \country{China}
}
\authornote{Work done during their internships at Tsinghua University.}

\author{Jingyan Xu}
\affiliation{
DCST, Tsinghua University\\
Beijing  \country{China}
}
\authornotemark[1]

\author{Ziyi Ye}
\affiliation{
Fudan University\\
Shanghai  \country{China}
}

\author{Qingyao Ai}
\authornote{Corresponding author}
\email{aiqy@tsinghua.edu.cn}
\affiliation{
DCST, Tsinghua University\\
Beijing  \country{China}
}

\author{Yiqun Liu}
\affiliation{
DCST, Tsinghua University\\
Beijing  \country{China}
}

\renewcommand{\shortauthors}{Su, et al.}

\begin{abstract}
Large Language Models (LLMs) frequently exhibit hallucinations, generating content that appears fluent and coherent but is factually incorrect.
Such errors undermine trust and hinder their adoption in real-world applications. 
To address this challenge, two distinct research paradigms have emerged: model-centric Hallucination Detection (HD) and text-centric Fact Verification (FV). 
Despite sharing the same goal, these paradigms have evolved in isolation, using distinct assumptions, datasets, and evaluation protocols. This separation has created a research schism that hinders their collective progress.
In this work, we take a decisive step toward bridging this divide. We introduce UniFact\footnote{We have open-sourced all the code, data, and baseline implementation at: \url{https://github.com/oneal2000/UniFact/}}, a unified evaluation framework that enables direct, instance-level comparison between FV and HD by dynamically generating model outputs and corresponding factuality labels. 
Through large-scale experiments across multiple LLM families and detection methods, we reveal three key findings: 
(1) No paradigm is universally superior; 
(2) HD and FV capture complementary facets of factual errors; 
and (3) hybrid approaches that integrate both methods consistently achieve state-of-the-art performance.
Beyond benchmarking, we provide the first in-depth analysis of why FV and HD diverged, as well as empirical evidence supporting the need for their unification. 
The comprehensive experimental results call for a new, integrated research agenda toward unifying Hallucination Detection and Fact Verification in LLMs.

\end{abstract}

\maketitle

\section{Introduction}
\label{sec-intro}

Large Language Models (LLMs) have fundamentally reshaped the digital landscape, powering next-generation search engines and automating complex tasks that previously seemed impossible~\cite{brown2020language,su2025surge,feng2024large,fourney2024magentic}. 
However, their deployment in high-stakes, real-world applications is critically hampered by a dangerous failure mode: the generation of fluent yet factually incorrect content, a phenomenon known as hallucination ~\cite{hallusurvey,su2024unsupervised,manakul2023selfcheckgpt,su2024mitigating}. 
In domains like healthcare~\cite{yang2023largemed,thirunavukarasu2023large,peng2023study}, legal~\cite{su2025judge,su2024stard,wang2024lekube}, and journalism, where factual reliability is a critical requirement, hallucinations can erode user trust, spread misinformation, and lead to harmful consequences. 
Therefore, developing robust mechanisms to detect these factual errors is not just a technical challenge but a foundational requirement for building trustworthy and reliable intelligent systems.

To address this critical challenge, the research community has developed two distinct paradigms: 
{Fact Verification} (FV)~\cite{thorne2018fever,hu2022dual,krishna2022proofver,lewis2020retrieval} and {Hallucination Detection} (HD)~\cite{hallusurvey,MIND,INSIDE,liu2025attentionguidedselfreflectionzeroshothallucination,hdndes,seu,malinin2020uncertainty,nie2018combining}.
Fact Verification is a well-established, text-centric research area rooted in information retrieval~\cite{chen2022gere,chen2022evidencenet,hu2023read} and fake news detection~\cite{shu2019beyond,zhou2020survey,oshikawa2020survey}. 
Typically, FV evaluates a statement's factuality based on external knowledge sources, such as web documents or databases. 
A defining characteristic of FV is its origin-agnostic nature: it evaluates a statement's factuality regardless of whether it is human-authored or machine-generated.
In this paradigm, the text is treated as an isolated, static artifact, where factuality is assessed based on the consistency between the claim and retrieved evidence.
In contrast, Hallucination Detection is a more recent paradigm that emerged alongside the rise of LLMs. 
The fundamental distinction lies in the object of evaluation: HD specifically targets content generated by LLMs. 
Unlike FV, which evaluates text in isolation, HD introduces a new dimension of introspective evaluation by leveraging the generative process itself.
While HD methods can also employ external knowledge retrieval similar to FV, the mainstream research focuses on model-centric signals, such as token-level uncertainty, activation dynamics, and consistency across sampled generations~\cite{MIND,INSIDE,manakul2023selfcheckgpt}. 
Consequently, HD is inherently model-centric, allowing it to identify hallucinations even in the absence of external evidence.

Despite sharing the same ultimate goal of identifying factual errors, these two research directions have evolved in surprising isolation, which is evidenced by their use of distinct benchmarks, evaluation metrics, and publication venues.
The root of this divergence lies in a fundamental incompatibility between evaluation paradigms. 
FV is inherently text-centric: its benchmarks, such as FEVER~\cite{thorne2018fever}, consist of claims treated as standalone texts, where their origin is not considered. 
HD methods, however, depend on internal model states that FV settings cannot provide.
Attempts to bridge this gap by constructing static benchmarks of LLM outputs (e.g., HELM~\cite{su2024unsupervised}) remain limited, as such datasets cannot evaluate new or out-of-distribution models, nor capture dynamic generative signals.
This incompatibility has led to the development of two parallel research tracks, resulting in redundant efforts and leaving fundamental questions unanswered.
It remains unclear whether one paradigm is inherently superior, how their strengths differ, or whether they can be effectively combined.
Moreover, their isolation prevents leveraging their complementary nature: HD’s internal model signals could augment FV’s evidence-based reasoning, while FV’s retrieved context could ground HD’s uncertainty signals.

To this end, we take a decisive step toward bridging this long-standing divide. 
We present the first large-scale, systematic empirical study to directly compare Hallucination Detection and Fact Verification under a single, unified setting. 
Rather than a mere benchmarking effort, our work provides a comprehensive investigation into the fundamental relationship between these paradigms.
This unified approach allows us, for the first time, to address three key research questions (RQs) that have remained unexplored due to the fields' historical separation:
\vspace{-0.5mm} 

\begin{enumerate}[leftmargin=*]

    \item \textbf{Overall Performance (RQ1)}: 
    How do Hallucination Detection and Fact Verification methods differ in performance when detecting factual errors on the same LLM outputs?
    
    \item \textbf{Complementary Strengths (RQ2)}: 
    Do HD and FV capture distinct aspects of factual inaccuracy, showing complementary strengths, or do they overlap and fail on similar errors?
    
    \item \textbf{Hybrid Performance (RQ3)}: Can the signals from both paradi-gms be effectively combined? Can such a hybrid detection system surpass either approach in isolation?
    
\end{enumerate}
\vspace{-0.5mm} 

\noindent This comparison was previously considered infeasible due to the paradigms’ fundamental incompatibility. Our work makes this comparison possible by introducing \textbf{UniFact}, a unified evaluation framework designed to overcome the fundamental incompatibility that has historically split the field. 
Unlike static benchmarks that rely on pre-generated, fixed LLM's outputs, UniFact adopts a dynamic evaluation process. It prompts any LLM to generate answers to factual questions in real time. The framework then employs an automated labeling system to determine correctness by comparing the generated output against ground-truth answers and their corresponding pre-labeled reference knowledge. This dynamic process is the key: it simultaneously yields both the textual outputs required for FV methods and the internal model signals needed for HD methods. This design enables the direct, instance-level, head-to-head comparison of both paradigms on the same generated content.

Our large-scale experiments, conducted via UniFact across diverse datasets, model families (e.g., LLaMA, Qwen), and detection methods, yield several crucial insights that directly answer our research questions. First, addressing RQ1, we find that no single paradigm is universally superior; performance varies significantly with the underlying LLM and the nature of the task. Second, and most critically (addressing RQ2), our analysis provides strong empirical evidence of complementarity. HD and FV methods consistently succeed on different subsets of factual errors, confirming they capture distinct yet synergistic dimensions of factuality. Finally (addressing RQ3), building on this discovery, we demonstrate that simple hybrid methods integrating both paradigms consistently and significantly outperform all individual approaches, establishing a new state-of-the-art in factual error detection.

The implications of these findings are twofold. For the research community, they motivate a paradigm shift from isolated studies toward a unified research agenda that embraces this synergy. For practitioners, our results offer a clear takeaway: robust factuality assessment requires combining internal, model-based signals with external, evidence-based reasoning.
In summary, this paper makes the following key contributions:

\begin{itemize}[leftmargin=*] 

\item \textbf{Analytical}: We identify and analyze the conceptual and methodological schism between Hallucination Detection and Fact Verification, highlighting its consequences for progress in AI factuality.

\item \textbf{Methodological}: We introduce UniFact, a unified evaluation framework. It overcomes this divide and enables direct, fair, and scalable comparison between HD and FV.

\item \textbf{Empirical}: Through the first large-scale comparative study, we provide strong evidence that HD and FV exhibit distinct yet complementary strengths.

\item \textbf{Practical}: We demonstrate that simple hybrid strategies combining both paradigms achieve new state-of-the-art performance.

\end{itemize}

\vspace{-1mm}

\section{Problem Formulation}
This section establishes formal definitions for Fact Verification (FV) and Hallucination Detection (HD). 
While these terms are often used interchangeably in recent literature, distinguishing them is crucial for understanding the evaluation landscape. 
As outlined in the introduction and illustrated in Figure~\ref{pic:compare}, the primary distinction lies in the object of evaluation: FV targets the semantic truthfulness of text in an origin-agnostic manner, whereas HD targets the factual correctness of the text generated by an LLM. 
We formally define HD and FV in \S~\ref{sec-def-fv} and \S~\ref{sec-def-hd}, respectively, and unify their evaluation metrics in \S~\ref{sec-def-eval}.

\subsection{Fact Verification}\label{sec-def-fv}
Fact Verification (FV) is defined as the task of assessing the factual correctness of a textual claim by grounding it in authoritative external evidence. 
A defining characteristic of FV is that it is {origin-agnostic} and {content-centric}: it evaluates the text as a static artifact, independent of whether it was authored by a human or generated by an LLM.
Formally, let $y$ denote a textual claim (e.g., a sentence or an atomic proposition), and $\mathcal{E}$ represent a set of trusted external evidence (e.g., retrieved documents, knowledge graph triples). 
FV treats the verification process as a classification problem between the claim and the evidence, disregarding the source that produced $y$. 
The task is formulated as estimating the probability that $y$ is {non-factual} given $\mathcal{E}$:

\begin{equation}
    p_{\text{FV}} = D_{\text{FV}}(y, \mathcal{E}) \in [0, 1],
\end{equation}

\noindent where $D_{\text{FV}}$ is a scoring function quantifying the factual error, such that $p_{\text{FV}} \approx 1$ indicates that $y$ conflicts with $\mathcal{E}$ (non-factual), while $p_{\text{FV}} \approx 0$ implies consistency. 
In this paradigm, the focus is exclusively on the alignment between the claim content and the world knowledge provided by $\mathcal{E}$~\cite{thorne2018fever,min2023factscore}.

\begin{figure}[t]
\centering
\includegraphics[width=\columnwidth]{Pic/compare2.pdf}

\caption{An illustration of comparison between Hallucination Detection (HD) and Fact Verification (FV). HD targets LLM outputs leveraging both intrinsic features and external knowledge (dashed arrow), whereas FV targets generic textual claims using exclusively retrieved evidence.}

\label{pic:compare}
\end{figure}

\subsection{Hallucination Detection}\label{sec-def-hd}
In contrast to FV, Hallucination Detection (HD) is {model-centric} and specifically targets the factual accuracy of the content generated by LLMs. 
HD aims to identify instances where the model generates content that is unfaithful to the source input or factually incorrect relative to world knowledge. 
Unlike FV, which evaluates text independently of its origin, HD is inherently model-centric and targets factual inconsistencies specifically in text generated by an LLM.
Formally, let $M$ be a language model that generates an output sequence $y$ given an input context $x$. HD estimates the probability that $y$ contains a hallucination. This estimation can be derived from two distinct sources of signals:

\begin{itemize}[leftmargin=*]
    \item \textbf{Intrinsic Signals ($\mathcal{F}_M$):} Features derived directly from the LLM's generative process, including predictive entropy, activation patterns, cross-sample consistency, etc.
    
    \item \textbf{Extrinsic Signals ($\mathcal{E}$):} External evidence retrieved to verify the generated content, comparable to FV, but utilized to evaluate the factual integrity of the LLM's response.
\end{itemize}

\noindent Therefore, the general formulation for HD is:
\begin{equation}
    p_{\text{HD}} = D_{\text{HD}}(y, x, \mathcal{F}_M, \mathcal{E}) \in [0, 1],
\end{equation}
\noindent Crucially, while external evidence $\mathcal{E}$ is a valid information source for HD, most existing Hallucination Detection approaches rarely incorporate $\mathcal{E}$, reflecting the paradigm’s emphasis on leveraging intrinsic signals $\mathcal{F}_M$ as its core methodological advantage. Consequently, HD enables introspective evaluation based solely on model-internal signals, avoiding the need for external retrieval.

\subsection{Evaluation}\label{sec-def-eval}

Despite their distinct methodologies, HD and FV can be unified under a single evaluation framework. 
Both tasks can be formulated as a binary classification problem. 
Specially, let $s_i \in [0,1]$ denote the predicted score for instance $i$, representing the probability that the text is non-factual (FV) or hallucinated (HD), and let $l_i^* \in \{0,1\}$ denote the ground-truth label, where $l_i^*=1$ signifies a negative sample. 
Standard classification metrics can be adopted to assess performance, specifically {Accuracy} and the Area Under the ROC Curve ({AUC}). AUC is particularly emphasized to evaluate the model's ranking capability independent of specific decision thresholds, addressing the potential class imbalance in evaluation datasets.

\section{Unified Evaluation Framework}
\label{sec-eval-frame}

In this section, we introduce our unified evaluation framework, designed to bridge the long-standing divide between Hallucination Detection and Fact Verification. We begin by examining why these two closely related paradigms have historically evolved in isolation, and we analyze the incompatibilities in existing benchmarks that have impeded cross-paradigm evaluation (§\ref{subsec-incompatibility}). We then present the core design principles underlying our framework (§\ref{subsec-principles}), followed by a detailed description of its architecture (§\ref{subsec-architecture}).

\subsection{The Benchmark Incompatibility Challenge}\label{subsec-incompatibility}
Before detailing our framework design, it is crucial to analyze the technical root causes that have historically segregated these two highly related paradigms. The divergence stems from a fundamental incompatibility between the \textit{text-centric} design of traditional FV datasets and the \textit{model-centric} requirements of HD methods. Traditional FV benchmarks, such as FEVER~\cite{thorne2018fever}, are inherently text-centric. They consist of static claim-label pairs that are agnostic to the origin of the claim. 
This structure renders them unsuitable for Hallucination Detection methods (e.g., SelfCheckGPT~\cite{manakul2023selfcheckgpt}, MIND~\cite{su2024unsupervised}), which fundamentally rely on analyzing the generation process, such as internal states or cross-sample consistency. 
These generative signals are absent in static FV datasets, creating an impassable barrier for unified evaluation.

A seemingly intuitive solution to bridge this gap is to construct static benchmarks comprised of pre-generated LLM outputs (e.g., HELM~\cite{su2024unsupervised}). While valuable, we argue that reliance on static collections of model outputs suffers from two prohibitive limitations in the rapidly evolving landscape of AI:
\begin{itemize}[leftmargin=*]
    \item \textbf{Model Obsolescence:} The pace of LLM development is extraordinary. A dataset constructed with outputs from today's state-of-the-art models (e.g., LLaMA-3) risks becoming an archival snapshot rather than a durable evaluation tool as new architectures emerge. 
    \item \textbf{Inability to Evaluate New Models:} Crucially, static benchmarks cannot evaluate HD methods on new or out-of-distribution (OOD) models. Since HD performance is often tightly coupled to specific model characteristics, a framework that cannot test methods on arbitrary new LLMs fundamentally restricts progress and fails to measure true generalizability.
\end{itemize}

To overcome these structural limitations, a paradigm shift is required from analyzing {static artifacts} to evaluating {dynamic processes}. An effective unified framework must not rely on fixed outputs but instead on a core set of instructions (questions) that can trigger generation from \textit{any} target LLM on the fly. This dynamic approach ensures that (1) FV methods receive the textual content they require, (2) HD methods gain access to the real-time internal states of the generating model, and (3) the evaluation remains perpetually relevant regardless of model evolution. This insight directly motivates the design of UniFact.

\subsection{Design Principles}
\label{subsec-principles}

To address the limitations of static benchmarks, our unified framework is built upon three foundational design principles that ensure a robust, equitable, and future-proof evaluation.

\begin{enumerate}[leftmargin=*]

\item \textbf{Model-Agnosticism via On-the-Fly Generation.} To combat model obsolescence, the framework abandons static datasets of outputs. Instead, it uses a core set of factual questions as the instruction to generate outputs dynamically from \textit{any} target LLM. This ensures the framework's perpetual relevance for evaluating methods on current and future LLMs.

\item \textbf{Equitable and Principled Comparison.} The framework is feasible for both hallucination detection and fact verification methods. It provides HD methods with access to the model's internal states during on-the-fly generation, while supplying FV methods with the generated text and an external corpus. Despite these different mechanisms, both are evaluated on the identical, ultimate task: predicting the factual correctness of a textual content (the output of that LLM).

\item \textbf{Scalability through Automated Labeling.} To move beyond the slow and subjective process of manual annotation, the framework employs an automatic labeling system. It automatically determines the factuality of any generated response by comparing it against the ground-truth answer and relevant knowledge, enabling large-scale, objective, and reproducible evaluations across any number of LLMs.

\end{enumerate}

\begin{figure}[t]
\centering
\includegraphics[width=\columnwidth]{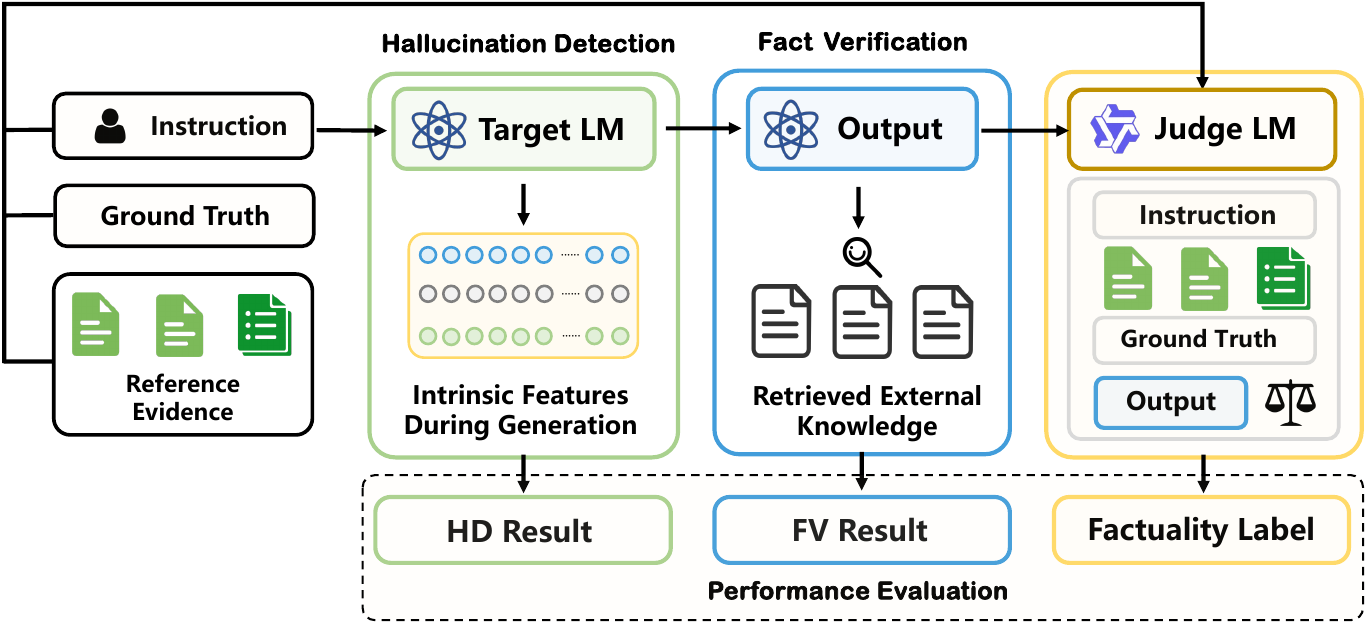}
\caption{An illustration of the UniFact evaluation framework. The pipeline generates answers dynamically and evaluates both hallucination detection and fact verification against a shared factuality label.}
\label{pic:framework}
\end{figure}

\subsection{Framework Architecture and Workflow}
\label{subsec-architecture}

As illustrated in Figure~\ref{pic:framework}, UniFact operates as a modular pipeline composed of three distinct stages: dynamic generation, automated annotation, and unified evaluation. This workflow transforms a static instruction set into a dynamic, reliably labeled benchmark that can evaluate both HD and FV paradigms.

\subsubsection{Stage 1: Dynamic Instance Generation}
The process initiates with an input triplet $(q, A^*, E^*)$, where $q$ is a factual question, $A^*$ represents the set of validated answers, and $E^*$ denotes authoritative evidence. 
Unlike static benchmarks, we treat $q$ as a prompt to trigger the target LLM $M_{\mathrm{target}}$ to produce a response $y_{\mathrm{gen}}$ in real-time.

To accommodate Hallucination Detection (HD) methods that require access to internal model states (white-box) or multiple stochastic samples (black-box), this stage facilitates the real-time extraction of all relevant generative signals during the decoding process. Consequently, it allows for the construction of a model-specific instance that encapsulates both the textual output $y_{\mathrm{gen}}$ and the requisite features, ensuring that downstream HD methods can obtain the necessary raw data for analysis.

\subsubsection{Stage 2: Reference-Based Automated Annotation}
To establish a ground-truth factuality label for the dynamically generated answer  $y_{\mathrm{gen}}$, we employ a reference-based verification procedure using an external evaluator model $M_{\mathrm{eval}}$. 
The evaluator receives the tuple $(q, y_{\mathrm{gen}}, A^*, E^*)$ and is tasked with a discriminative judgment: determining whether $y_{\mathrm{gen}}$ is semantically consistent with the provided ground truth $A^*$ and evidence $E^*$.
This process yields a rigorously annotated evaluation instance, formally denoted as $(q, y_{\mathrm{gen}}, l^*)$, where $l^*$ represents the binary factuality label (Consistent/Inconsistent).

It is important to emphasize that the role of $M_{\mathrm{eval}}$ is fundamentally distinct from that of $M_{\mathrm{target}}$. While the target model engages in open-ended generation, $M_{\mathrm{eval}}$ operates under a fixed, rubric-governed judgment protocol. The rubric specifies precise criteria for factual agreement, contradiction, partial support, and unsupported assertions, thereby enforcing a controlled and reproducible decision-making process. Crucially, $M_{\mathrm{eval}}$ is granted privileged access to the ground-truth answer and evidence, allowing it to anchor its judgment in the authoritative reference materials rather than relying on its own parametric knowledge. As a result, the task reduces to a constrained consistency-checking problem rather than a knowledge-generation problem, substantially narrowing the space for evaluator hallucination and minimizing model subjectivity.

This evidence-rich setup also provides strong empirical reliability. As detailed in Section~\ref{subsec:human}, human annotators exhibit 97.42\% agreement with $M_{\mathrm{eval}}$ on hallucination labels and 99.02\% agreement on non-hallucination labels. These results confirm that, under a well-specified rubric and with explicit access to $(A^*, E^*)$, the evaluator model produces factuality labels $l^*$ that closely track human judgments and are sufficiently dependable for large-scale automated annotation.

\subsubsection{Stage 3: Unified Evaluation Interface}
The final stage bridges the paradigm gap by evaluating HD and FV methods on the identical prediction target $l^*$, despite their differing input requirements.
\begin{itemize}[leftmargin=*]
    \item \textbf{Hallucination Detection (HD):} These methods receive the generated text $y_{\mathrm{gen}}$ along with the intrinsic model features captured in Stage 1. They are evaluated on their ability to detect non-factual content using only self-contained signals.
    \item \textbf{Fact Verification (FV):} These methods treat $y_{\mathrm{gen}}$ as a claim to be verified. Crucially, to simulate a realistic verification scenario, FV methods are \textit{not} given access to the ground-truth evidence $E^*$. Instead, they must retrieve supporting documents from an external corpus (e.g., Wikipedia) to predict the label.
\end{itemize}
This design ensures a fair yet distinct comparison: both paradigms aim to predict the same ground truth $l^*$, but HD is tested on its utilization of internal model uncertainty, while FV is tested on its ability to retrieve and reason over external knowledge.

\subsection{Implementation Details}

This section outlines key implementation details of our framework, covering dataset construction, judge model configuration, and the human verification procedure.

\subsubsection{Dataset Construction}

The foundation of our framework is a static set of diverse instructions, which we constructed by sourcing questions from five factual QA datasets. 
As our paradigm requires an automated system to verify the factual correctness of generated outputs, we deliberately selected datasets where each instance provides the three essential components for this process: a question (serving as the instruction), a ground-truth answer set, and relevant knowledge in the form of supporting relevant documents (enabling reliable factuality labeling).
The five chosen datasets fulfill these criteria and were selected to cover a spectrum of hallucination challenges. TriviaQA (TQA)~\cite{triviaqa}, NQ-Open (NQ)~\cite{nq}, and PopQA (PQA)~\cite{mallen-etal-2023-trust} are open-domain benchmarks that assess factual recall across diverse generations. Meanwhile, 2WikiMultihopQA (2WQA)~\cite{ho2020constructing} evaluates multi-hop reasoning, for which we use the Bridge and Comparison question types (denoted as Bridge and Comp). HotpotQA (HQA)~\cite{yang2018hotpotqa} also targets multi-hop reasoning, from which we select Comparison-type questions (referred to as HComp). 
For each dataset, we randomly sample 500 questions as test instances, ensuring a balanced and statistically meaningful evaluation. 
For the external knowledge base, we utilize the Wikipedia dumps processed by DPR~\cite{karpukhin2020dense}\footnote{https://github.com/facebookresearch/DPR/tree/main} as the external corpus.

\subsubsection{Judge Model Configuration}

For the LLM-as-Judge component ($M_{\text{eval}}$), we employ the Qwen-2.5-32B model~\cite{yang2024qwen2}, a high-capacity LLM demonstrating strong instruction-following capabilities. 
Specifically, we utilize a carefully constructed prompt (see Appendix~\ref{app-prompt}) that explicitly instructs $M_{\text{eval}}$ to identify factual discrepancies based strictly on the provided ground-truth answer set $A^*$ and evidence $E^*$. 
This configuration ensures that the judge performs a reference-based verification rather than utilizing its internal parametric memory, thereby aligning the automated evaluation with the rubric-defined objective.

\subsubsection{Human Verification}
\label{subsec:human}
To validate the reliability of the UniFact framework, we conducted a manual verification of the labels produced by the automatic labeling system illustrated in Figure~\ref{pic:framework}. 
The human-annotation process is conducted as follows. First, we filtered out instances where the target LLM refused to answer (e.g., responses containing "I'm unable to xxx" or "I am not sure"). 
This resulted in a set of 1,602 generated answers, of which the LLM-as-Judge identified 890 as containing hallucinations and 712 as factually correct. 
We then tasked four undergraduate students to verify these labels manually. 
The human annotators agreed with the judge's "hallucination" label in 97.42\% of cases and with the "non-hallucination" label in 99.02\% of cases. This high level of agreement confirms the reliability of our auto-labeling approach. 
It is crucial to note that this high accuracy is achieved because the automatic labeling system is provided with the ground truth answer ($A^*$) as well as the evidence ($E^*$) from the dataset. This contrasts with the real-world scenario for FV methods, which must retrieve their own, often imperfect, evidence from an external corpus.

\section{Unified Empirical Study}

In this section, we conduct a systematic empirical study to bridge the gap between Hallucination Detection and Fact Verification. We first introduce our study design and experimental setup. We then present a detailed analysis of three proposed research questions, exploring the comparative performance, complementary strengths, and integration potential of these two paradigms.

\subsection{Study Design}
\label{sec-overview}

To bridge the historical gap between Hallucination Detection and Fact Verification, we conduct a large-scale empirical study designed to systematically compare these two paradigms within a unified framework. Our investigation is structured to provide clear and direct answers to the following three research questions (RQs):

\begin{itemize}[leftmargin=*]
\item \textbf{RQ1: A Head-to-Head Performance Comparison.} How do SOTA Hallucination Detection and Fact Verification methods perform when evaluated on the same LLM-generated content? Are there consistent winners?
\item \textbf{RQ2: Uncovering Complementary Strengths.} Do these two paradigms possess unique, complementary strengths in identifying different types of factual errors, or do their capabilities largely overlap?
\item \textbf{RQ3: Performance of Hybrid Methods.} Can we create a more robust and accurate system by combining signals from both paradigms, outperforming any single approach?
\end{itemize}

To answer these questions, we organize our empirical study into three logical and progressive steps. 
First, to answer \textbf{RQ1}, we implement a wide array of classic and SOTA baselines from both paradigms and benchmark their performance head-to-head on a unified testbed. 
This provides the first direct comparison of their relative effectiveness.
Second, building on these initial results, we quantitatively analyze the synergy between the two paradigms to address \textbf{RQ2}. We demonstrate that the mutual enhancement achieved by integrating HD and FV significantly surpasses the gains from combining methods within a single paradigm.
Furthermore, we conduct a detailed case study in Section~\ref{sec-casestudy} to add qualitative depth to our analysis of {RQ2}. 
Finally, motivated by the strong evidence of their synergy, we investigate \textbf{RQ3} by designing and evaluating two simple yet effective hybrid methods. We test fusion at both the final probability score level and the system pipeline level, showing that even straightforward integration can significantly boost overall performance.

\begin{table*}
\caption{Performance of Hallucination Detection and Fact Verification methods under our proposed unified evaluation framework. The best results in each column are highlighted in bold, and the second-best results are underlined. The reported metrics in the table are Area Under the Curve (AUC) scores.}
\label{table-main}
\centering
\setlength\tabcolsep{3pt}
  {
    \begin{tabular}{ll ccc ccc ccc ccc}
      \toprule
      \multicolumn{2}{c}{} & \multicolumn{6}{c}{\textbf{Meta-Llama-3.1-8B-Instruct}} & \multicolumn{6}{c}{\textbf{Qwen2.5-14B-Instruct}} \\
      \cmidrule(lr){3-8} \cmidrule(lr){9-14}
      \multicolumn{2}{c}{} & \textbf{Bridge} & \textbf{Comp} & \textbf{HComp} & \textbf{NQ} & \textbf{PQA} & \textbf{TQA} & \textbf{Bridge} & \textbf{Comp} & \textbf{HComp} & \textbf{NQ} & \textbf{PQA} & \textbf{TQA} \\
      \midrule
      \multirow{13}{*}{\textbf{HD}}
      & \textbf{LNPP} & 0.6557 & 0.6950 & 0.7259 & 0.7361 & 0.6103 & 0.7476 & 0.5278 & 0.7189 & 0.7062 & 0.6466 & 0.6129 & 0.6912 \\
      & \textbf{LNPE} & 0.6991 & 0.7648 & 0.7540 & 0.7370 & 0.5853 & 0.7524 & 0.4953 & \underline{0.7389} & 0.7211 & 0.6820 & 0.6242 & 0.7022 \\
      & \textbf{SCG-MQA} & 0.4785 & 0.6799 & 0.6841 & 0.7082 & 0.6675 & 0.7290 & 0.5351 & 0.6294 & 0.6386 & 0.6129 & 0.6026 & 0.7291 \\
      & \textbf{SCG-NG} & 0.3598 & 0.3291 & 0.4152 & 0.6553 & 0.6649 & 0.7109 & 0.4266 & 0.5559 & 0.5424 & 0.6012 & 0.6701 & 0.6524 \\
      & \textbf{SCG-BS} & 0.4097 & 0.4194 & 0.6229 & 0.7631 & 0.5750 & 0.7672 & 0.5333 & 0.5566 & 0.6674 & 0.6844 & 0.6541 & 0.6942 \\
      & \textbf{SCG-NLI} & 0.5082 & 0.5870 & 0.7010 & 0.7149 & 0.5513 & 0.7537 & 0.5701 & 0.6209 & 0.6722 & 0.6401 & 0.4790 & 0.7031 \\
      & \textbf{SAPLMA} & 0.4335 & 0.4714 & 0.6451 & 0.5758 & 0.7356 & 0.6013 & 0.5365 & 0.4368 & 0.5969 & 0.5357 & 0.5505 & 0.6164 \\
      & \textbf{EUBHD} & 0.5904 & 0.6442 & 0.6591 & 0.6714 & 0.7012 & 0.6613 & 0.4546 & 0.5607 & 0.7154 & 0.6836 & \textbf{0.8130} & 0.6639 \\
      & \textbf{MIND} & 0.5675 & 0.6898 & 0.7296 & 0.6098 & 0.8043 & 0.7472 & \underline{0.7214} & 0.4790 & 0.6126 & 0.6232 & 0.5560 & 0.6242 \\
      & \textbf{PTrue} & 0.5185 & 0.5027 & 0.4403 & 0.5593 & 0.2880 & 0.3944 & 0.5400 & \textbf{0.7446} & 0.6965 & 0.7048 & 0.6239 & 0.7325 \\
      & \textbf{SE} & 0.4446 & 0.4851 & 0.4260 & 0.5436 & 0.3141 & 0.5280 & 0.5317 & 0.5519 & 0.5340 & 0.5195 & 0.4339 & 0.5351 \\
      & \textbf{SEU} & 0.5136 & 0.6669 & 0.7085 & 0.7212 & 0.5623 & 0.7508 & 0.5291 & 0.5914 & 0.7284 & 0.7019 & 0.5533 & 0.7449 \\
      & \textbf{SIndex} & 0.3991 & 0.5299 & 0.6004 & 0.6938 & 0.3156 & 0.6594 & 0.4959 & 0.5939 & 0.6501 & 0.6716 & 0.4311 & 0.7128 \\
      \midrule
      \multirow{4}{*}{\textbf{FV}}
      & \textbf{LLM-Q} & 0.5034 & 0.5162 & 0.6456 & 0.7405 & 0.5080 & 0.6954 & 0.5901 & 0.5840 & 0.7392 & 0.7086 & 0.6963 & 0.7907 \\
      & \textbf{BERT-Q} & 0.6485 & 0.7607 & 0.7409 & 0.6154 & \underline{0.8256} & 0.7215 & 0.4946 & 0.6399 & 0.7099 & 0.5956 & 0.7617 & 0.7019 \\
      & \textbf{LLM-QA} & 0.5354 & 0.5707 & 0.6597 & 0.7537 & 0.4874 & 0.7019 & 0.5054 & 0.5881 & \underline{0.7691} & 0.7474 & 0.6831 & 0.8120 \\
      & \textbf{BERT-QA} & 0.6362 & 0.7813 & 0.7393 & 0.6165 & 0.8141 & 0.7364 & 0.5086 & 0.6367 & 0.6986 & 0.6290 & 0.7518 & 0.7338 \\
      \midrule
      \multirow{5}{*}{\textbf{Unify}}
      & \textbf{BERT-Q+EUBHD} & 0.6414 & 0.8126 & 0.7612 & 0.6846 & \textbf{0.8552} & 0.7485 & 0.4688 & 0.6391 & 0.7614 & 0.6609 & 0.8074 & 0.7394 \\
      & \textbf{BERT-Q+LNPE} & \textbf{0.7268} & \underline{0.8128} & \underline{0.7902} & 0.7062 & 0.7902 & \underline{0.7904} & 0.4845 & 0.7285 & 0.7668 & 0.6901 & 0.7830 & 0.7563 \\
      & \textbf{BERT-QA+MIND} & 0.5738 & 0.7669 & 0.7572 & 0.6300 & 0.8063 & 0.7692 & \textbf{0.7239} & 0.5726 & 0.6801 & 0.6500 & 0.6839 & 0.6904 \\
      & \textbf{LLM-QA+LNPE} & 0.6369 & 0.6903 & 0.7540 & \textbf{0.8222} & 0.5392 & 0.7902 & 0.5004 & 0.6687 & \textbf{0.8230} & \textbf{0.8028} & 0.7062 & \textbf{0.8597} \\
    & \textbf{Pipeline} &\underline{0.7026} &\textbf{0.8176} &\textbf{0.7989} &\underline{0.7877} & 0.8089 &\textbf{0.8252} &0.4958 &0.7239 &0.7491 &\underline{0.7567} &\underline{0.8098} &\underline{0.8175} \\                                  

      \bottomrule
    \end{tabular}
  }
\end{table*}

\subsection{Experimental Setup}
\label{sec-setup}

\subsubsection{Backbone Models and Implementation Details.} 
All experiments are conducted using open-source pre-trained large language models (LLMs). We select LLMs of varying sizes and architectural series, including \textbf{Qwen2.5-14B-Instruct}~\citep{yang2024qwen2} and \textbf{LLaMA-3.1-8B-Instruct}~\citep{Llama-3.2-1B-Instruct}. All models are deployed via their official Hugging Face implementations. 
We adopt the default hyperparameters and chat templates provided in the official repositories, with the only modification being the use of greedy decoding to ensure reproducibility of the results. 
All experiments are conducted on NVIDIA A100 GPUs with 40GB VRAM.

\subsubsection{Hallucination Detection Baselines.} 
We evaluate the following set of hallucination detection baselines.
\textbf{Semantic Entropy (SE)}~\cite{se} computes token-level entropy and applies clustering to refine uncertainty estimates. 
\textbf{Semantic Embedding Uncertainty (SEU)}~\cite{seu} captures response variability through pairwise embedding similarity analysis. 
\textbf{SINdex}~\cite{sindex} extends SE by modeling both intra-group and inter-group inconsistencies.
\textbf{Length-Normalized Predictive Entropy (LNPE)}~\cite{malinin2020uncertainty} estimates predictive entropy and normalizes it by sequence length.
\textbf{PTrue}~\cite{ptrue} estimates the probability that a generation is truthful using token-level likelihoods. 
\textbf{Length-Normalized Predictive Probability (LNPP)}~\cite{manakul2023selfcheckgpt} evaluates hallucination likelihood based on length-normalized token probabilities.
\textbf{SAPLMA}~\cite{azaria2023internal} trains a classifier that detects hallucination based on the LLM's activation values. 
\textbf{MIND}~\cite{su2024unsupervised} detects an LLM's hallucination based on the final layer's token embeddings.
\textbf{EUBHD}~\cite{zhang2023enhancing} represents a state-of-the-art uncertainty-based method with enhanced predictive distribution modeling. 
Finally, we evaluate the \textbf{SelfCheckGPT (SCG)}~\cite{manakul2023selfcheckgpt} family, which detects hallucinations based on sampling consistency. We consider four specific variants: SCG-BS, SCG-MQA, SCG-NLI, and SCG-NG.

\subsubsection{Fact Verification Baselines} 

To provide a comprehensive comparison, we implement baselines representing the two dominant paradigms in automated fact-checking: LLM-based Verification~\cite{min2023factscore,wei2024long}) and NLI-based Verification~\cite{thorne2018fever,soleimani2020bert}). To ensure a fair comparison within our unified framework, we standardize the retrieval module across all methods, using BM25~\cite{robertson2009probabilistic} for evidence retrieval from the DPR Wikipedia corpus~\cite{karpukhin2020dense}.
Implementation details for all baselines are provided in Appendix~\ref{appendix-fv}.

\begin{itemize}[leftmargin=*] 

\item \textbf{LLM-based Verification.} Recent works such as FActScore~\cite{min2023factscore} and SAFE~\cite{wei2024long} have demonstrated the effectiveness of using LLMs to verify claims against retrieved evidence. Following this paradigm, we implement two variants based on their query formulation strategies: 

\begin{itemize}[leftmargin=*] 

\item \textbf{LLM-Q:} This baseline adopts the standard "Retrieve-then-Verify" workflow. Given a question, we retrieve evidence passages using the question as the query. An evaluator LLM is then prompted to verify whether the target LLM's generated response is supported by the retrieved evidence. 

\item \textbf{LLM-QA:} This variant retrieves evidence using both the question {and} the generated answer. This allows the verifier to capture evidence relevant to claims made in the response. The same evaluator LLM is used for the final verification. 

\end{itemize}

\item \textbf{NLI Verification.} We also evaluate the classical "Retriever-Reader" architecture, which remains a strong baseline in fact-checking benchmarks like FEVER~\cite{thorne2018fever}. Adapting the architecture from \citet{soleimani2020bert}, we implement two variants that utilize a fine-tuned BERT classifier as the verifier, distinguished by their retrieval strategies:

\begin{itemize}[leftmargin=*]

\item \textbf{BERT-Q:} This baseline retrieves evidence passages using only the question as the query. The retrieved passages are then concatenated with the question and answer to form the input for a BERT-based classifier, which predicts whether the answer contains factual errors.

\item \textbf{BERT-QA:} This variant retrieves evidence using both the question and the generated answer. Similar to LLM-QA, this strategy aims to capture evidence that is semantically closer to the specific claims in the response. The same BERT-based classifier is employed to predict the verification label based on the retrieved context.

\end{itemize}

\end{itemize}

\paragraph{Implementation Note on Standardization.} 
While these baselines are conceptually grounded in SOTA fact verification frameworks like FActScore~\cite{min2023factscore} and BERT-based verification systems~\cite{soleimani2020bert}, we adapt their implementations to fit a unified evaluation setting. 
Specifically, we standardize the retrieval backbone (BM25) and the knowledge corpus (Wikipedia) across all experiments. 
This control eliminates potential bias introduced by different retrieval systems or proprietary databases used in original implementations. 
Consequently, to distinguish our standardized adaptations from the specific system configurations in prior work, we denote them using descriptive names (e.g., LLM-QA, BERT-Q) rather than their original names (e.g., FActScore).

\subsection{RQ1: A Head-to-Head Performance Analysis}\label{sec-rq1}

The results presented in Table~\ref{table-main} reveal a distinct performance divergence between Hallucination Detection (HD) and Fact Verification (FV), indicating that neither paradigm offers a universally superior solution. We observe that the performance of HD methods is unstable and depends largely on the specific LLM being evaluated. 
For instance, methods that achieve high scores on the LLaMA family often see a significant drop when applied to Qwen models. This instability suggests that reliance on internal signals (e.g., predictive entropy) for hallucination detection may not transfer reliably across different model families. 
In contrast, FV methods demonstrate greater stability. Notably, incorporating the generated answer into retrieval queries (the QA-based strategy) consistently outperforms question-only retrieval across most datasets. This confirms that for FV, the primary bottleneck is finding the right evidence to match the claim, rather than the model architecture itself.

Beyond the stability trends, a fine-grained comparison of individual datasets further confirms that neither paradigm maintains a strict dominance. As detailed in Table~\ref{table-main}, the "winner" fluctuates significantly depending on the dataset and model combination. For instance, on Meta-Llama-3.1-8B-Instruct, HD methods secure the top performance on four out of six datasets, whereas FV methods are superior on two out of six datasets (e.g., PQA and Comp). However, this advantage disappears on Qwen2.5-14B-Instruct, where the two paradigms split the victories evenly. 
This lack of a consistent winner strongly suggests that the two paradigms are sensitive to different types of factual errors. Rather than being redundant, their performance divergence implies a potential for orthogonality, which directly motivates our quantitative analysis of their complementarity in the following section (RQ2). Note that the performance of hybrid methods (the ``Unify'' block) is distinct from this head-to-head comparison and will be discussed in detail under RQ3.

\subsection{Paradigm Synergy Analysis (RQ2)}
\label{sec:quantitative_synergy}

To quantitatively address \textbf{RQ2}, we conduct a comparative analysis of method synergy within and across HD and FV paradigms. Specifically, we investigate whether combining methods from different paradigms (i.e., HD+FV) yields greater complementarity than combining methods within the same paradigm (i.e., HD+HD or FV+FV).
To this end, we use three complementary metrics to evaluate the synergy between an arbitrary pair of methods, where the methods are denoted as $M_A$ and $M_B$. Let $C_A$ and $C_B$ be the sets of samples correctly classified by $M_A$ and $M_B$, respectively (regardless of whether they belong to the same or different paradigms), and let $W_A$ and $W_B$ be the corresponding sets of incorrectly classified samples. The total set of samples is $S$.
We formalize the synergy between any two methods using the following three metrics:

\begin{itemize}[leftmargin=*]
    \item \textbf{Average Complementarity Score (ACS)~\cite{kuncheva2003measures}.} This metric measures the proportion of samples correctly classified by exactly one of the two methods, quantifying their degree of non-overlapping correctness. Equivalently, ACS corresponds to the classical disagreement rate between two classifiers.
    
    \begin{align}
        \text{ACS}(M_A, M_B) = \frac{|(C_A \cap W_B) \cup (W_A \cap C_B)|}{|S|}.
    \end{align}

    \item \textbf{Average Synergy Gain (ASG)~\cite{shipp2002relationships}.} This metric captures the performance improvement of an ideal oracle ensemble over the best-performing individual method. It reveals the potential accuracy gain achievable in principle by combining $M_A$ and $M_B$.
    \begin{align}
        \text{ASG}(M_A, M_B) = \frac{|C_A \cup C_B|}{|S|} - \max\left(\frac{|C_A|}{|S|}, \frac{|C_B|}{|S|}\right).
    \end{align}
    In other words, ASG is the fraction of samples that are correctly classified by the oracle ensemble but misclassified by the better of the two individual methods.
    
    \item \textbf{Average Error Correction Rate (AECR)~\cite{kuncheva2003measures}.} 
This metric quantifies the capacity of one method to compensate for another's errors. We define the directional error correction rate of method $M_A$ for $M_B$, denoted as $R(A\ \mathrm{for}\ B)$, as the conditional probability that $M_A$ is correct given $M_B$ has failed:
\begin{align}
    R(A\ \mathrm{for}\ B) &= P(C_A \mid W_B) = \frac{|C_A \cap W_B|}{|W_B|} ,\\
    R(B\ \mathrm{for}\ A) &= P(C_B \mid W_A) = \frac{|C_B \cap W_A|}{|W_A|} .
\end{align}
The AECR is the symmetrical average of these two directional rates\footnote{AECR is applied only to pairs where both methods make at least one error (i.e., $|W_A|,|W_B|>0$), which holds for all methods in our experiments.}:
\begin{equation}
    \text{AECR}(M_A, M_B) = \frac{1}{2} \left( R(A\ \mathrm{for}\ B) + R(B\ \mathrm{for}\ A). \right)
\end{equation}

\end{itemize}

\begin{table}[t]
\centering
\caption{Synergy and complementarity analysis between and within paradigms. Cross-paradigm pairings consistently outperform intra-paradigm comparisons across all metrics, demonstrating strong complementary strengths.}
\label{tab:synergy_metrics}
\begin{tabular}{l|ccc}
\toprule
\textbf{Method Pairing}      & \textbf{ACS} & \textbf{ASG} & \textbf{AECR} \\ \midrule
\textbf{Intra-HD Average}             & 0.315        & 0.118        & 0.503         \\
\textbf{Intra-FV Average}             & 0.379        & 0.102        & 0.496         \\ \midrule
\textbf{Cross-Paradigm Average} & \textbf{0.428} & \textbf{0.144} & \textbf{0.634} \\ \bottomrule
\end{tabular}
\end{table}

For each of the three metrics, we compute the value for all unordered pairs of methods within the HD paradigm, within the FV paradigm, and across both paradigms. We then report the average over method pairs in each group. The resulting scores are presented in Table~\ref{tab:synergy_metrics}, providing quantitative evidence that the HD and FV paradigms possess complementary strengths. The notably higher ACS for cross-paradigm pairings indicates that their successes are more mutually exclusive than those of methods within the same paradigm, suggesting that they excel on different subsets of the problem space. 
Similarly, the superior ASG score reveals that combining HD and FV methods offers the greatest potential for performance improvement, exceeding any intra-paradigm combination.
Most importantly, the AECR exhibits a substantial increase for cross-paradigm pairings. This implies that the failure modes of the two paradigms are largely distinct: when one paradigm makes an error, the other is considerably more likely to be correct. 
Overall, this analysis supports the view that HD and FV are not merely different but genuinely complementary, providing an empirical basis for the development of hybrid models that integrate both paradigms.

\subsection{Case Study}
\label{sec-casestudy}

While the previous section established the statistical complementarity between HD and FV, this section investigates the underlying causes of their divergent performance. Through a detailed case study, we examine specific scenarios where each paradigm tends to fail. We highlight two distinct failure patterns: Fact Verification is fundamentally limited by its reliance on retrieval quality, often leading to errors when relevant evidence is missing. Conversely, Hallucination Detection is prone to false alarms caused by semantic flexibility, where the model's diverse phrasing is misinterpreted as hallucination. Analyzing these patterns helps clarify the nature of their synergy and motivates the design of the hybrid integration methods proposed in the subsequent section.

\subsubsection{Fact Verification} 
Our analysis suggests that the effectiveness of FV methods is fundamentally limited by the performance of the retrieval module and the availability of relevant external knowledge. 
As demonstrated in Table~\ref{tab:fv_case_studies}, this reliance on retrieval creates a critical vulnerability when the system fails to provide necessary evidence. Specifically, when the retriever cannot locate relevant passages (for instance, due to keyword mismatch in the LLM-Q setting), the verification module is compelled to make judgments without sufficient context. This absence of evidence results in unstable model behavior: the model may either show unfounded skepticism (leading to false alarms, as in Case 1) or accept false claims without verification (leading to missed errors, as in Case 2). In contrast, when the retrieval process is augmented with the generated answer (LLM-QA), the system is more likely to obtain relevant supporting documents, thereby allowing the FV model to make more robust predictions.

\subsubsection{Hallucination Detection}
Conversely, HD methods are prone to {false alarms} triggered by the inherent {semantic flexibility} of language generation. 
Table~\ref{tab:hd_flexibility_cases} reveals a clear dichotomy: HD methods perform robustly on deterministic answers (Group A) but degrade significantly on expressive responses (Group B). In the latter case, a model may be factually confident about an event (e.g., the formation year of a band) yet lexically uncertain about how to articulate it. Current HD methods struggle to disentangle this valid semantic diversity from actual falsehoods. As a result, the high entropy caused by diverse phrasing is misinterpreted as hallucination, causing the detector to flag correct generations solely because they are phrased with variety.

\subsubsection{Implications for Hybrid Design}
Our qualitative analysis reveals that HD and FV operate with orthogonal error boundaries, providing the theoretical grounding for their integration. 
FV functions as a precision-oriented anchor: when it successfully retrieves evidence, its judgment is highly reliable, but it performs poorly when the retrieval module fails to fetch relevant evidence. 
Conversely, HD functions as a recall-oriented safety net: it can capture potential errors even without external knowledge, but suffers from noise due to semantic flexibility. 
Therefore, an optimal hybrid strategy should not assign equal importance to both. Instead, it should prioritize FV's judgment when retrieved evidence is relevant, while relying on HD's intrinsic signals when the retrieval result is not relevant.
This conditional complementarity directly motivates the hierarchical design of our proposed hybrid framework in RQ3.

\begin{table}[t]
    \caption{Failure Analysis of Fact Verification (FV). FV judgments are unstable and heavily dependent on retrieval quality. When retrieval fails to surface relevant evidence, FV models act erratically, leading to both False Alarms (Case 1) and Missed Detections (Case 2).}
    \label{tab:fv_case_studies}
    \centering
    \footnotesize
    \renewcommand{\arraystretch}{1.2}
    \setlength{\tabcolsep}{4pt}
    
    \begin{tabularx}{\linewidth}{p{1.8cm} X}
    \toprule
    \multicolumn{2}{c}{\textbf{Impact of Retrieval Quality on Verification Verdicts}} \\
    \midrule
    
    \multicolumn{2}{l}{\cellcolor{gray!10}\textit{\textbf{Case 1: Retrieval Failure $\rightarrow$ False Alarm (Unjustified Skepticism)}}} \vspace{2pt} \\
    \textbf{Question} & Who plays Cindy Lou Who on the Grinch? \\
    \textbf{LLM Output} & The role of Cindy Lou Who in the 2000 live-action film "How the Grinch Stole Christmas" is played by \textbf{Taylor Momsen}. \\
    \textbf{Ground Truth} & Taylor Michel Momsen \hfill \textit{(Label: Factual)} \\
    \cmidrule(l{2pt}r{2pt}){1-2}
    
    \textbf{State: Retrieval \newline Failure} & 
    \textbf{Evidence:} \textcolor{gray}{\textit{[No relevant passage regarding Taylor Momsen]}} \newline
    \textbf{Verdict:} \textcolor{red}{\textbf{Hallucination}} \hfill \textit{(Error: False Alarm)} \\
    
    \cmidrule[0.5pt](l{2pt}r{2pt}){2-2} 
    
    \textbf{State: Retrieval \newline Success} & 
    \textbf{Evidence:} \textit{\textcolor{blue}{Cindy Lou Who} is a generous young girl who was introduced in the book "How the Grinch Stole Christmas!" In the 2000 live action film, "How the Grinch Stole Christmas!" she is played by actress \textcolor{blue}{Taylor Momsen}.} \newline
    \textbf{Verdict:} \textcolor{green!40!black}{\textbf{Factual}} \hfill \textit{(Correct)} \\
    
    \midrule
    
    \multicolumn{2}{l}{\cellcolor{gray!10}\textit{\textbf{Case 2: Retrieval Failure $\rightarrow$ Missed Detection (Blind Confirmation)}}} \vspace{2pt} \\
    \textbf{Question} & Where did the theme song from SWAT come from? \\
    \textbf{LLM Output} & The theme song from the popular 1970s-80s TV show "S.W.A.T." was composed by \textbf{Michel Legrand and Al Burton}. \\
    \textbf{Ground Truth} & Written by Barry De Vorzon \hfill \textit{(Label: Hallucination)} \\
    \cmidrule(l{2pt}r{2pt}){1-2}
    
    \textbf{State: Retrieval \newline Failure} & 
    \textbf{Evidence:} \textcolor{gray}{\textit{[No relevant passage regarding composer]}} \newline
    \textbf{Verdict:} \textcolor{red}{\textbf{Factual}} \hfill \textit{(Error: Missed Detection)} \\
    
    \cmidrule[0.5pt](l{2pt}r{2pt}){2-2}
    
    \textbf{State: Retrieval \newline Success} & 
    \textbf{Evidence:} \textit{\textcolor{blue}{Theme from "S.W.A.T."} is an instrumental song written by \textcolor{blue}{Barry De Vorzon} and performed by American funk group Rhythm Heritage, released on their debut album "Disco-Fied".} \newline
    \textbf{Verdict:} \textcolor{green!40!black}{\textbf{Hallucination}} \hfill \textit{(Correct)} \\
    
    \bottomrule
    \end{tabularx}
\end{table}

\begin{table}[t]
    \caption{\textbf{Failure Analysis of Hallucination Detection (HD).} HD methods struggle to distinguish \textit{Factual Uncertainty} from \textit{Lexical Uncertainty}. While they excel at verifying deterministic facts (Group A), they frequently generate False Alarms on semantically flexible responses (Group B).}
    \label{tab:hd_flexibility_cases}
    \centering
    \footnotesize
    \renewcommand{\arraystretch}{1.15}
    \setlength{\tabcolsep}{3pt}
    
    \begin{tabularx}{\linewidth}{p{0.8cm} X}
    \toprule
    \multicolumn{2}{c}{\textbf{Impact of Semantic Flexibility on HD Verdicts}} \\
    \midrule
    
    \multicolumn{2}{p{\dimexpr\linewidth-2\tabcolsep}}{\cellcolor{blue!5}\textbf{Group A: Deterministic Facts (Low Lexical Entropy)}} \\
    \multicolumn{2}{p{\dimexpr\linewidth-2\tabcolsep}}{\scriptsize{\textit{Characteristic: Single, concise answers. Low token-level uncertainty.}}} \vspace{2pt} \\
    \midrule
    
    \textbf{Q1} & How many states are in the United States? \\
    \textbf{LLM} & There are 50 states in the United States. \\
    \textbf{Result} & \textbf{\textcolor{green!40!black}{Correct Verdict}} (Low Uncertainty $\rightarrow$ Factual) \\
    \cmidrule(l{2pt}r{2pt}){1-2}
    
    \textbf{Q2} & What is the water cycle also known as? \\
    \textbf{LLM} & The water cycle is also known as the hydrologic cycle. \\
    \textbf{Result} & \textbf{\textcolor{green!40!black}{Correct Verdict}} (Low Uncertainty $\rightarrow$ Factual) \\
    
    \midrule
    \midrule
    
    \multicolumn{2}{p{\dimexpr\linewidth-2\tabcolsep}}{\cellcolor{red!5}\textbf{Group B: Semantically Flexible Responses (High Lexical Entropy)}} \\
    \multicolumn{2}{p{\dimexpr\linewidth-2\tabcolsep}}{\scriptsize{\textit{Characteristic: Valid facts expressed with high variation (word choice/order), causing false spikes in uncertainty metrics.}}} \vspace{2pt} \\
    \midrule
    
    \textbf{Q3} & Which band formed first, Awolnation or Foo Fighters? \\
    \textbf{LLM} & \textbf{Foo Fighters formed first.} Foo Fighters were formed in 1994 by Nirvana's drummer Dave Grohl, following the band's dissolution... \\
    \textbf{Truth} & Foo Fighters formed first. \\
    \textbf{Result} & \textbf{\textcolor{red}{False Alarm}} (High Uncertainty due to phrasing $\rightarrow$ Hallucination) \\
    \cmidrule(l{2pt}r{2pt}){1-2}
    
    \textbf{Q4} & Who directed the classic 30s western Stagecoach? \\
    \textbf{LLM} & \textbf{The classic 1939 western 'Stagecoach' was directed by John Ford.} It is considered one of the greatest Westerns ever made... \\
    \textbf{Truth} & John Ford. \\
    \textbf{Result} & \textbf{\textcolor{red}{False Alarm}} (High Uncertainty due to phrasing $\rightarrow$ Hallucination) \\
    
    \bottomrule
    \end{tabularx}
\end{table}

\subsection{RQ3: Investigation of Hybrid Strategies}
\label{sec-rq3}

Building on the findings in RQ2, which demonstrated that Hallucination Detection (HD) and Fact Verification (FV) exhibit statistical complementarity and mechanistically distinct failure modes, we now address \textbf{RQ3}: {Can we operationalize this synergy into high-performance hybrid systems?}
To this end, we propose and evaluate two integration paradigms: a \textbf{Score-Level Fusion} approach and an \textbf{Evidence-Aware Pipeline} designed to address the specific structural limitations of FV identified in our case study.

\subsubsection{Hybrid Method Design}

\paragraph{Strategy I: Score-Level Fusion.}
Since HD and FV capture different aspects of hallucination, merging their signals can smooth out individual errors. 
We propose a straightforward linear combination of their normalized scores. 
Let $S_{\text{HD}}$ and $S_{\text{FV}}$ be the probability scores from a Hallucination Detection method (e.g., LNPE) and a Fact Verification method (e.g., LLM-QA), respectively. 
{To ensure consistency}, we map both scores to represent the likelihood of a hallucination (i.e., higher scores indicate lower quality).
The hybrid score is calculated as:
\begin{equation}
    S_{\text{Hybrid}} = \lambda S_{\text{HD}} + (1-\lambda) S_{\text{FV}}.
\end{equation}
In our experiments, we set $\lambda = 0.5$, effectively taking the average of the two signals. 
We adopt this unweighted setting to demonstrate that the hybrid method is robust and effective even without dataset-specific hyperparameter tuning.
We test pairings such as \textbf{LLM-QA + LNPE} to verify these gains.

\paragraph{Strategy II: Evidence-Aware Pipeline.} 
While Strategy I enhances robustness through score-level fusion, it does not explicitly address the failure modes identified in Section~\ref{sec-casestudy}. 
Specifically, FV systems are reliable when evidence is explicitly retrieved (yielding ``Supported'' or ``Contradicted'' verdicts) but become unreliable when relevant evidence is absent (i.e., ``Not Enough Information'' or NEI). 
To address this limitation, we design an \textit{Evidence-Aware Pipeline} that prioritizes external grounding, falling back to internal signals only when retrieval fails:
\begin{enumerate}[leftmargin=*] 
    \item \textbf{Step 1: Retrieval \& Verification.} An FV module (specifically {LLM-QA}) retrieves evidence and classifies the claim--evidence relationship as \textit{Supported}, \textit{Contradicted}, or \textit{NEI}. 
    \item \textbf{Step 2: Conditional Branching.} We employ a cascading decision mechanism. If the FV yields a definitive prediction (\textit{Supported} or \textit{Contradicted}), the system directly accepts this result. Conversely, if the output is \textit{NEI}, the system acknowledges the lack of external support. Instead of forcing a potentially erroneous verification, the pipeline bypasses the FV and falls back to an HD method (e.g., \textbf{LNPE}), deriving the final score solely from the model's internal uncertainty signals. 
\end{enumerate} 
This conditional design effectively mitigates the limitations of FV when retrieval results are not relevant. Crucially, decoupling retrieval dependence from internal signals prevents the propagation of errors caused by low-quality evidence retrieval, ensuring robust detection even when external knowledge is unavailable.

\subsubsection{Performance Analysis}

The experimental results in Table~\ref{table-main} validate the efficacy of integrating model-centric and data-centric paradigms. A comprehensive examination of the ``Unify'' block against individual baselines reveals that hybrid strategies consistently establish a new upper bound for performance across diverse datasets and model architectures. This consistent superiority confirms that internal uncertainty signals and external evidence retrieval operate in orthogonal feature spaces.
Regarding the specific strategies, the Score-Level Fusion approach demonstrates that a simple linear combination can effectively mitigate the high variance inherent in single-paradigm methods. By aggregating divergent signals, this strategy smoothens the noise from individual prediction errors, yielding a more stable evaluation that outperforms standalone HD or FV methods in most scenarios. 
Furthermore, the Evidence-Aware Pipeline exhibits an even more distinct advantage, frequently achieving the optimal performance among all compared methods. 
This performance leap can be attributed to the pipeline's conditional mechanism, which addresses the fundamental ``blind spot'' of retrieval-based verification. 
In instances where external knowledge is insufficient (the NEI cases), forcing a verification verdict often introduces noise; the pipeline avoids this by dynamically retreating to internal uncertainty signals, which remain a reliable proxy for truthfulness in the absence of evidence.

Notably, hybrid methods mitigate the performance instability observed in RQ1. Unlike standalone HD methods, which exhibit fluctuations between LLaMA and Qwen families, hybrid approaches maintain robustness across diverse backbones. This suggests that combined signals effectively compensate for the miscalibration of internal confidence inherent in standalone HD. Consequently, the proposed frameworks not only enhance accuracy but also offer a generalized solution independent of model-specific variations.

\section{Related Work}

This section reviews existing work in Fact Verification (FV) and Hallucination Detection (HD), highlighting the lack of unified benchmarks that motivates our proposed \textbf{UniFact} framework.

\subsection{Fact Verification}

\subsubsection{Traditional Pipeline-based Methods}

The establishment of the FEVER benchmark~\cite{thorne2018fever} formalized FV as a multi-stage pipeline consisting of document retrieval, evidence selection, and textual entailment (NLI). Early approaches, such as the baseline provided by ~\citet{thorne2018fever}, relied on TF-IDF retrieval and simple classifiers, suffering from error propagation between stages. 
Subsequent works addressed these limitations through end-to-end neural architectures. For instance, NSMN~\cite{nie2019combining} proposed a unified neural semantic matching network for joint retrieval and verification, while GEAR~\cite{zhou2019gear} introduced graph-based reasoning to model dependencies across multiple evidence sentences. 
With the advent of pre-trained language models, BERT-based approaches~\cite{soleimani2020bert} further improved performance by enhancing contextual representations for both retrieval and claim verification. Extensions like FEVEROUS~\cite{aly2021feverous} and SciFact~\cite{wadden2020fact} expanded this paradigm to structured data (tables) and scientific domains, respectively.

\subsubsection{Fact Verification in the Era of LLMs}

With the rapid advancement of LLMs in natural language processing, an increasing number of studies explore their application to fact verification. Techniques such as retrieval-augmented generation (RAG~\cite{su2024dragin,su2025parametric,dong2025decoupling,tu2025rbft,su2025dynamic,susigir,wang2025knowledge}) and few-shot in-context learning (ICL~\cite{brown2020languagemodelsfewshotlearners,dong2024survey,wang2025decoupling}) enable LLMs to perform accurate truth assessments and generate supporting evidence by incorporating relevant contextual knowledge.
These methods typically use a retrieval module (either a simple retriever~\citep{zhai2008statistical,su2023caseformer,robertson2009probabilistic,su2023wikiformer,ma2023caseencoder,fang2024scaling,su2025pre} or a more sophisticated retrieval system~\citep{su2023thuir2,salemi2024towards,chen2022web,tu2022reinforcement,tu2025generalized}) to identify evidence related to the statement from knowledge sources and then provide both the statement and retrieved evidence as input to the LLM, guiding it to make judgments through ICL. 
For instance, ~\citet{ssinghal2024evidence} proposes a method that combines evidence extraction via retrieval with truth prediction using ICL with LLMs. ~\citet{deng2024cram} further investigates how to improve the reliability of RAG systems by reordering retrieved documents based on credibility, thereby mitigating the impact of erroneous information on LLM predictions.
Furthermore, ~\citet{zhang2023towards} designs a hierarchical prompting method that guides LLMs to decompose complex statements into more manageable sub-statements and verify each progressively, aiming to improve the robustness of fact-checking complex claims.

\subsection{Hallucination Detection}

\subsubsection{White-Box Methods: Leveraging Internal States}

White-box methods require access to the model's parameters and activations~\cite{wang2025joint,su2024unsupervised}. Early works focused on token-level probability distributions, using metrics like perplexity or entropy to quantify uncertainty~\cite{malinin2020uncertainty}. 
For instance, {EUBHD}~\cite{zhang2023enhancing} propose an uncertainty-based hallucination detection method based on predictive distribution modeling. 
More recent approaches leverage deeper internal representations. SAPLMA~\cite{SAPLMA} and MIND~\cite{MIND} train supervised classifiers based on the model's hidden states to detect hallucinations. INSIDE~\cite{INSIDE} analyzes the covariance matrix of internal states across multiple sampled outputs to detect inconsistencies. 
From another angle, HD-NDEs~\cite{hdndes} employs neural differential equations to model the dynamic evolution of hidden states.

\subsubsection{Black-Box Methods: Behavioral Consistency}

For closed-source models where internal states are inaccessible, black-box methods rely on sampling-based consistency checks. The prevailing hypothesis is that if an LLM "knows" a fact, it will generate consistent answers across multiple stochastic samples; if it is hallucinating, the answers will diverge. SelfCheckGPT~\cite{manakul2023selfcheckgpt} is a representative approach that samples multiple responses and measures their consistency using NLI or LLM-based prompting. Other methods, such as SEU~\cite{seu}, quantify this consistency via semantic embeddings, while InterrogateLLM~\cite{yehuda-etal-2024-interrogatellm} reconstructs the query from the answer to check for semantic drift.
While HD methods excel at capturing the model's intrinsic uncertainty, they often lack external grounding. A model can be "confidently wrong" (low uncertainty, high consistency) due to mimicked falsehoods in pre-training data, a failure mode that internal signals alone often fail to detect.

\subsection{The Evaluation Gap}

A fundamental barrier to unifying these two paradigms lies in the incompatibility of existing benchmarks.
Traditional FV benchmarks like FEVER~\cite{thorne2018fever} consist of static claim-evidence pairs. While suitable for text-based verification, they are unusable for HD methods.
Conversely, benchmarks designed for LLM evaluation, such as TruthfulQA~\cite{lin2022truthfulqa} or HaluEval~\cite{li2023halueval}, often rely on pre-generated text or fixed multiple-choice formats. These static benchmarks suffer from rapid obsolescence; they cannot evaluate HD methods on new architectures (e.g., LLaMA-3 vs. LLaMA-2) nor facilitate the extraction of real-time internal signals required for white-box detection.
This dichotomy has forced researchers to evaluate HD and FV in isolation. To address this, our framework, \textbf{UniFact}, introduces a dynamic evaluation paradigm. By generating responses on-the-fly and automatically verifying them against external references, UniFact simultaneously provides the internal signals required by HD and the textual evidence required by FV, enabling the first rigorous head-to-head comparison of these complementary approaches.

\section{Conclusion}

In this work, we bridge the historical gap between Hallucination Detection (HD) and Fact Verification (FV) by introducing {UniFact}, a unified dynamic factuality evaluation framework for LLMs. Through the first direct comparison of these two paradigms, we reveal that they represent orthogonal dimensions of factuality assessment rather than redundant approaches. 
Specifically, while FV provides robust grounding when external evidence is available, HD leverages internal uncertainty to serve as a crucial safeguard when retrieval fails. 
Capitalizing on this synergy, we demonstrate that integrating these distinct signals establishes a new state-of-the-art in factual error detection. Ultimately, this paper advocates for a paradigm shift from isolated research tracks toward unified factuality assessment. We hope UniFact serves as a foundation for future research, paving the way for trustworthy AI systems that seamlessly integrate internal introspection with external knowledge.

\clearpage

\bibliographystyle{ACM-Reference-Format}
\bibliography{sample-base}

\clearpage

\appendix

\section{Prompt Templates}

This appendix provides the complete prompt templates used in our unified evaluation framework.

\subsection{LLM-as-Judge Prompt}
\label{app-prompt}

The LLM-as-Judge is a critical component for dynamic ground truth annotation. We use the following prompt template when both ground truth answers and passages are available:

\begin{tcolorbox}[colback=gray!5, colframe=gray!50, boxrule=0.5pt, arc=2mm]
\small
\textbf{Prompt:}
\par\vspace{0.3em}
You are an expert evaluator assessing the factual accuracy of a Generated Answer based *only* on the provided Golden Answer and Golden Passages for a given Question. Determine if the Generated Answer introduces significant factual errors, hallucinations, or contradictions compared to the trusted information.
\par\vspace{0.5em}
[Instruction]\\
Evaluate the Generated Answer based *strictly* on the Golden Answer and Golden Passages. Do not use external knowledge. Identify if there are clear factual inaccuracies or information present in the Generated Answer that is not supported by or contradicts the provided golden information.
\par\vspace{0.5em}
[Question]\\
\{question\}
\par\vspace{0.5em}
[Golden Answer]\\
\{golden\_answer\}
\par\vspace{0.5em}
[Golden Passages]\\
\{golden\_passages\_formatted\}
\par\vspace{0.5em}
[Generated Answer]\\
\{generated\_answer\}
\par\vspace{0.5em}
[Evaluation Task]\\
Choose the option that best describes the Generated Answer's factuality based *only* on the provided Golden Answer and Golden Passages:\\
(A) Accurate: The Generated Answer is factually consistent with the Golden Answer and Golden Passages. It does not contain significant inaccuracies or unsupported claims.\\
(B) Inaccurate/Hallucinated: The Generated Answer contains significant factual errors, contradictions, or information not supported by the Golden Answer and Golden Passages.
\par\vspace{0.5em}
Provide your final choice by writing only the letter (A or B).\\
Choice:
\end{tcolorbox}
When ground truth passages are unavailable, we use a simplified version that evaluates the generated answer solely against the golden answer, omitting the [Golden Passages] section.

\subsection{Fact Verification Prompt (LLM-based)}

For LLM-based fact verification methods (LLM-Q and LLM-QA), we employ the following prompt to assess the hallucination level of generated answers against retrieved evidence:

\begin{tcolorbox}[colback=gray!5, colframe=gray!50, boxrule=0.5pt, arc=2mm]
\small
\textbf{Prompt:}
\par\vspace{0.3em}
Based on the following information, determine if the generated answer contains hallucinations (inaccurate or fabricated information):
\par\vspace{0.5em}
Question: \{question\}
\par\vspace{0.5em}
Generated Answer: \{answer\}
\par\vspace{0.5em}
Reference Documents:\\
\{retrieved\_passages\}
\par\vspace{0.5em}
Please carefully analyze whether the generated answer is consistent with the information in the reference documents. If the answer contains information not mentioned in or contradicting the reference documents, it should be considered as containing hallucinations.
\par\vspace{0.5em}
Rate the hallucination level on a scale from 0 to 1:\\
- 0: Completely reliable (no hallucination)\\
- 1: Completely unreliable (severe hallucination)
\par\vspace{0.5em}
Output only a number between 0 and 1, no explanation needed.
\end{tcolorbox}

\section{Implementation Details and Hyperparameters}
\label{app-detail}

\subsection{Answer Generation and Hallucination Detection}

The main answer is generated using sampling-based decoding with the following hyperparameters:
\begin{itemize}[leftmargin=*]
    \item \textbf{Max New Tokens:} 30
    \item \textbf{Temperature:} 0.8
    \item \textbf{Top-p:} 0.9
\end{itemize}

For methods requiring multiple samples (e.g., SelfCheckGPT, Semantic Entropy, SEU, SIndex, PTrue), we generate 5 additional samples using the same hyperparameters. All hallucination detection methods follow their original hyperparameters as reported in their respective papers.

\subsection{Fact Verification Methods}
\label{appendix-fv}

\subsubsection{Evidence Retrieval with BM25}
All fact verification methods rely on BM25 retrieval from a Wikipedia corpus. We use the preprocessed Wikipedia dump provided by DPR~\cite{karpukhin2020dense}, which consists of 21 million passages, each containing approximately 100 words. The corpus is indexed using Elasticsearch with default BM25 parameters (k1=1.2, b=0.75). For each query, we retrieve the top-3 most relevant passages based on BM25 scoring.

We implement two retrieval strategies:
\begin{itemize}[leftmargin=*]
    \item \textbf{Question-Only Retrieval:} Uses only the original question as the search query. This approach is query-agnostic and does not depend on the generated answer.

    \item \textbf{Question+Answer Retrieval:} Concatenates the question and the generated answer to form the search query. This often yields more targeted evidence passages that are directly relevant to the specific claims in the generated answer.
\end{itemize}

Both retrieval strategies use the same Elasticsearch backend with BM25 scoring. Retrieved passages are then used as evidence for both LLM-based and BERT-based fact verification methods.

\subsubsection{LLM-based Fact Verification}
We implement two LLM-based fact verification methods that differ only in their retrieval strategy:
\begin{itemize}[leftmargin=*]
    \item \textbf{LLM-Q:} Retrieves passages using the question only, then uses an LLM (Qwen-2.5-32B) to verify the answer against retrieved evidence. The generation uses greedy decoding.

    \item \textbf{LLM-QA:} Similar to LLM-Q but retrieves passages using both question and answer as the query.
\end{itemize}

\subsubsection{BERT-based Fact Verification}
We employ a BERT-based classifier initialized with \texttt{bert-base-uncased} to predict whether a generated answer constitutes a hallucination. We train two variants that differ only in the retrieval query used to obtain supporting passages: \textbf{BERT-Q} retrieves passages using the question alone, while \textbf{BERT-QA} retrieves passages using the concatenated question and answer.

\paragraph{Input Representation.}
For a question $q$, a generated answer $a$, and a set of retrieved passages $\{p_1, p_2, \ldots, p_k\}$, we construct the input sequence by first concatenating the retrieved passages, then appending the question and answer as separate components. Formally, let $P = p_1 \oplus p_2 \oplus \cdots \oplus p_k$ denote the concatenated passage text, where $\oplus$ represents string concatenation with space delimiters. The complete input representation is:
\begin{equation}
    X(q, a, P) = [\mathrm{CLS}]\, P\, [\mathrm{SEP}]\, q\, [\mathrm{SEP}]\, a\, [\mathrm{SEP}],
\end{equation}
where $[\mathrm{CLS}]$ serves as the classification token and $[\mathrm{SEP}]$ tokens separate the passage, question, and answer components. This structure enables the model to jointly encode the retrieved evidence and the question-answer pair for verification.

\paragraph{Model Architecture.}
We feed the tokenized input $X(q, a, P)$ into the BERT encoder to obtain contextualized embeddings. The classifier extracts the final-layer $[\mathrm{CLS}]$ token embedding and applies a linear transformation for binary classification:
\begin{equation}
    \mathrm{h}_{\mathrm{CLS}} = \mathrm{BERT}_{[\mathrm{CLS}]}(X(q, a, P)),
\end{equation}
\begin{equation}
    \mathbf{z} = \mathrm{Dropout}(\mathrm{h}_{\mathrm{CLS}};\, p=0.2),
\end{equation}
\begin{equation}
    \mathrm{logits} = \mathbf{W}\mathbf{z} + \mathbf{b},
\end{equation}
where $\mathrm{h}_{\mathrm{CLS}} \in \mathbb{R}^{768}$ is the contextualized embedding of the $[\mathrm{CLS}]$ token, and $\mathbf{W} \in \mathbb{R}^{2 \times 768}$, $\mathbf{b} \in \mathbb{R}^{2}$ are learned parameters. The final prediction probability for the hallucination class is obtained via softmax normalization.

The model is optimized using the cross-entropy loss:
\begin{equation}
\mathcal{L} = -\frac{1}{N}\sum_{i=1}^{N}\left[y_i \log(\hat{p}_i) + (1-y_i)\log(1-\hat{p}_i)\right],
\end{equation}
where $\hat{p}_i = \mathrm{softmax}(\mathrm{logits}_i)_1$ is the predicted probability of hallucination, $y_i \in \{0,1\}$ denotes the ground truth label ($0$ for accurate, $1$ for hallucinated), and $N$ is the batch size.

\paragraph{Training Configuration.}
Both BERT classifiers are trained for 5 epochs with a learning rate of $2 \times 10^{-5}$ and batch size of 16, using the AdamW optimizer with linear warmup (10\% of total steps) and gradient clipping (max norm 1.0). Training data is automatically generated using our unified evaluation framework by collecting LLM-generated answers with corresponding LLM-as-Judge labels, and we use a disjoint slice of questions from those reserved for evaluation to prevent leakage. We perform a stratified 90/10 train-validation split and select the model checkpoint with the highest AUROC on the validation set.

\end{document}